\documentclass[10pt,twocolumn,letterpaper]{article}

\usepackage[pagenumbers]{cvpr} 

\usepackage{graphicx}
\usepackage{amsmath}
\usepackage{xcolor}
\usepackage{colortbl}
\usepackage{amssymb}
\usepackage{booktabs}

\usepackage[pagebackref,breaklinks,colorlinks]{hyperref}

\usepackage[capitalize]{cleveref}
\crefname{section}{Sec.}{Secs.}
\Crefname{section}{Section}{Sections}
\Crefname{table}{Table}{Tables}
\crefname{table}{Tab.}{Tabs.}

\usepackage{bbding}
\usepackage{xcolor}
\definecolor{green}{RGB}{0,255,0}
\definecolor{red}{RGB}{255,0,0}
\usepackage{multirow}
\usepackage{stfloats}
\usepackage{bm}
\def\cvprPaperID{8383} 
\def\confName{CVPR}
\def\confYear{2023}

\begin{document}

\title{JoyHallo: Digital human model for Mandarin}

\author{
    Sheng Shi,
    Xuyang Cao,
    Jun Zhao,
    Guoxin Wang\\
    JD Health International Inc. \\
}
\maketitle

\begin{abstract}
  In audio-driven video generation, creating Mandarin videos presents significant challenges. Collecting comprehensive Mandarin datasets is difficult, and the complex lip movements in Mandarin further complicate model training compared to English. In this study, we collected 29 hours of Mandarin speech video from JD Health International Inc. employees, resulting in the jdh-Hallo dataset. This dataset includes a diverse range of ages and speaking styles, encompassing both conversational and specialized medical topics. To adapt the JoyHallo model for Mandarin, we employed the Chinese wav2vec2 model for audio feature embedding. A semi-decoupled structure is proposed to capture inter-feature relationships among lip, expression, and pose features. This integration not only improves information utilization efficiency but also accelerates inference speed by 14.3\%. Notably, JoyHallo maintains its strong ability to generate English videos, demonstrating excellent cross-language generation capabilities.
  The code and models are available at \url{https://jdh-algo.github.io/JoyHallo/}. 
\end{abstract}

\section{Methods}
    \subsection{Overview}
        Artificial Intelligence Generated Content (AIGC) is the most promising and popular research in deep learning. Video generation tasks involve multimodal information such as audio and images, and present numerous challenges. Essential components of video generative models include the audio encoder (wav2vec\cite{wav2vecc2}), image encoder (VAE\cite{vae}), Transformer modules\cite{attention}, and diffusion frameworks\cite{diffusion}. Utilizing these models, AnimateAnyone\cite{animateanyone} proposed a face reenactment method that employs facial landmarks from a driving video to control the pose of a given source image while maintaining the identity of the source image.
        
        Hallo\cite{xu2024hallo} introduced an innovative audio-driven generative digital human model, significantly advancing video generation tasks. However, these models have primarily been trained and evaluated on English datasets. While they achieve satisfactory results in English video generation, they often underperform in Mandarin. The development of Mandarin digital human models has been constrained by two main factors: the lack of high-quality Mandarin datasets and the more complex lip movements in Mandarin compared to English.

\begin{figure*}[!htb]
        \centering
        \includegraphics[width=0.95\linewidth]{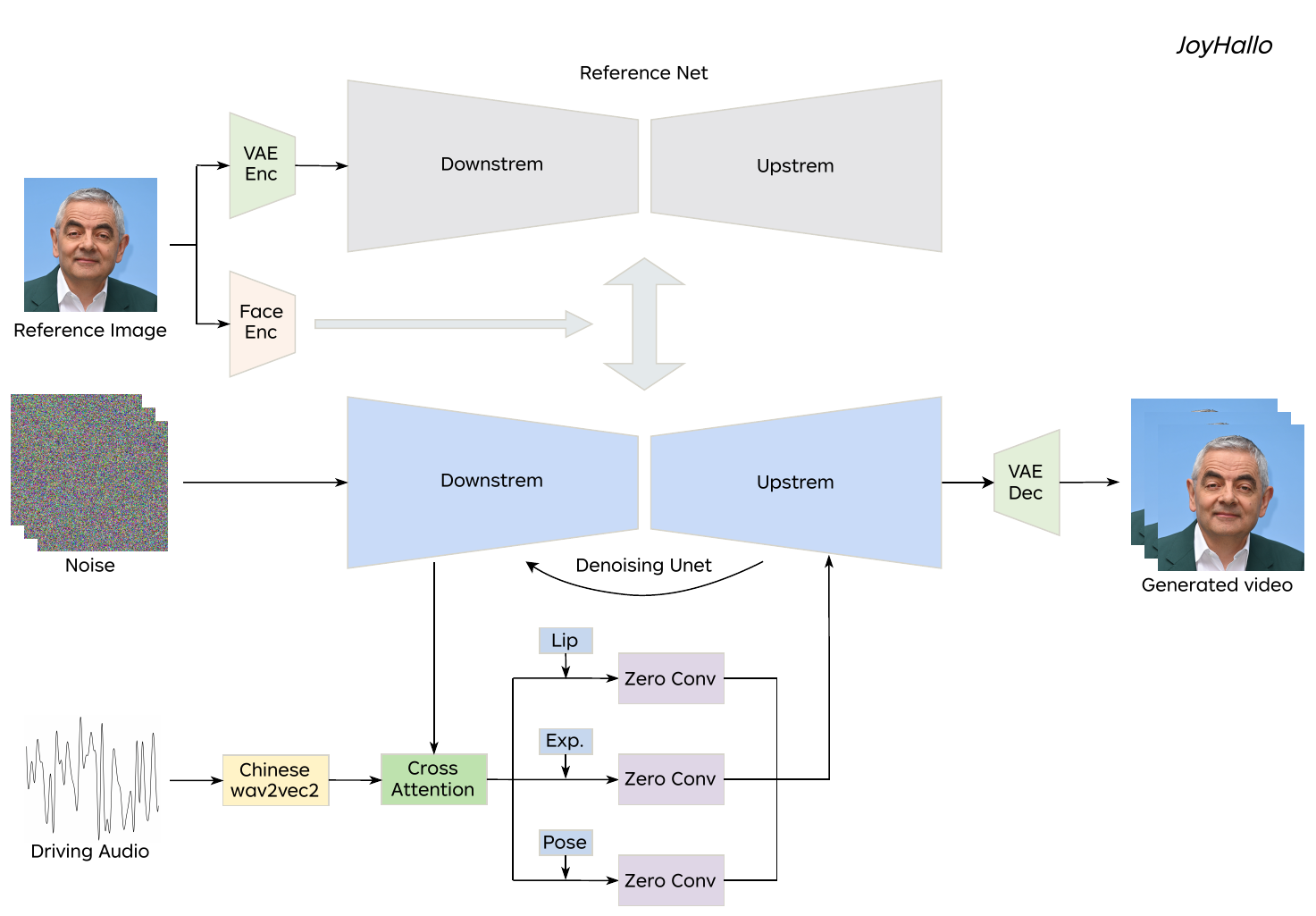}
        \caption{The framework of our realistic talking face generation model JoyHallo with semi-decoupled structure.}
        \label{fig:1}
        \end{figure*}

        As illustrated in Figure \ref{fig:1}, we introduce the JoyHallo model, which utilizes a semi-decoupled structure to address inadequacies in lip movement prediction.
        
    \subsection{Semi-decoupled Structure}
        The semi-decoupled structure is an innovative module designed to improve the accuracy of lip movement prediction in audio-driven video generation, particularly for Mandarin. This structure integrates and then separates key components involved in facial animation, enabling more precise modeling of complex lip movements and expressions.

        In traditional models, features such as lip movements, expressions, and head poses are often entangled, leading to suboptimal results due to the intricate nature of Mandarin phonetics. Some methods, like Hallo, employ a fully decoupled approach where facial features are completely separated and learned independently. While effective in isolating individual features, this approach may hinder the model's ability to learn hidden correlations between facial features.  With the complexity of facial features increases, it is challenging to maintain generation accuracy.

        The semi-decoupled structure initially couples these features and processes them together to capture their correlative characteristics using a Cross Attention module. Subsequently, a decoupling module is employed to separate different features. This separation allows the model to focus on the specific nuances of each feature without interference from others. The overall semi-decoupled module can be expressed as:
        \begin{equation}
        \begin{aligned}
        F_{coup} = CrossAtten(z_{t}, E_{audio}[A])
        \end{aligned}
        \end{equation}

        \begin{equation}
        \begin{aligned}
        F_{decoup} &= Conv_{zero}[F_{coup}*M_{Lip}]\\&+Conv_{zero}[F_{coup}*M_{Exp}]\\&+Conv_{zero}[F_{coup}*M_{Pose}]
        \end{aligned}
        \end{equation}
        where $z_{t}$ is a noisy latent variable of the reference image, and $E_{audio}[A]$ is an embedded feature of the reference audio. $M_{Lip}$, $M_{Exp}$, and $M_{Pose}$ are masks for the lip, expression, and pose, respectively.
        
        The semi-decoupled structure effectively balances independent feature processing with the ability to capture inter-feature relationships. Thereby enhancing overall performance and adaptability. By utilizing this structure, the JoyHallo model achieves greater accuracy in lip-sync and facial expressions, particularly in Mandarin video generation. Furthermore, the Chinese wav2vec2 \cite{chwav2vec} model is employed to enhance performance in Mandarin. This method not only increases the realism of the generated videos but also improves inference efficiency.

\section{Datesets}
\begin{figure}[!htb]
    \centering
    \includegraphics[width=1\linewidth]{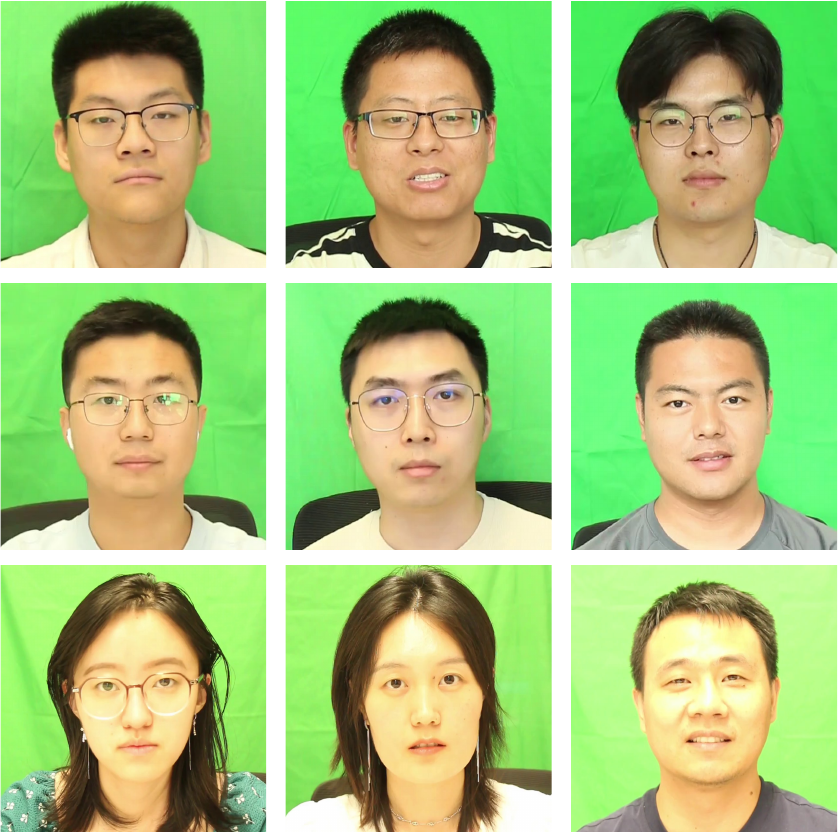}
    \caption{Talking faces in jdh-Hallo dataset.}
    \label{fig:2}
    \end{figure}

    As previously mentioned, the development of Mandarin talking faces encounters two major challenges. In addition to the complexity of Mandarin lip movements compared to English, there is also a lack of high-quality Mandarin datasets. Most existing datasets are in English or other languages. To address this issue, we conducted data collection with employees at JD Health International Inc., obtaining 29 hours of effective Mandarin speech videos. The data spans multiple age groups and various speaking styles, including daily conversational content and specialized medical topics. We processed this data with the Hallo \cite{xu2024hallo} model to create the Mandarin dataset jdh-Hallo. Figure \ref{fig:2} illustrates some examples from the jdh-Hallo dataset.
    
\section{Experiments}
    To validate the effectiveness of our method, we conducted extensive experiments focusing on video generation quality and speed. For video quality assessment, we employed several metrics: IQA\cite{wu2023qalign}, VQA\cite{wu2023qalign}, Sync-C\cite{syncnect}, Sync-D\cite{syncnect}, Smooth\cite{vbench}, Subject\cite{vbench}, and Background\cite{vbench}. IQA and VQA were used to evaluate image and video quality, respectively. Sync-C and Sync-D assessed lip synchronization in the generated videos. Smooth, Subject, and Background evaluated motion smoothness, subject consistency, and background consistency, respectively. For inference speed, we used memory usage and inference time as evaluation metrics. We used 8 NVIDIA A100 GPUs to train the JoyHallo model and made other experiments.

\begin{figure}[!htb]
    \centering
    \includegraphics[width=1\linewidth]{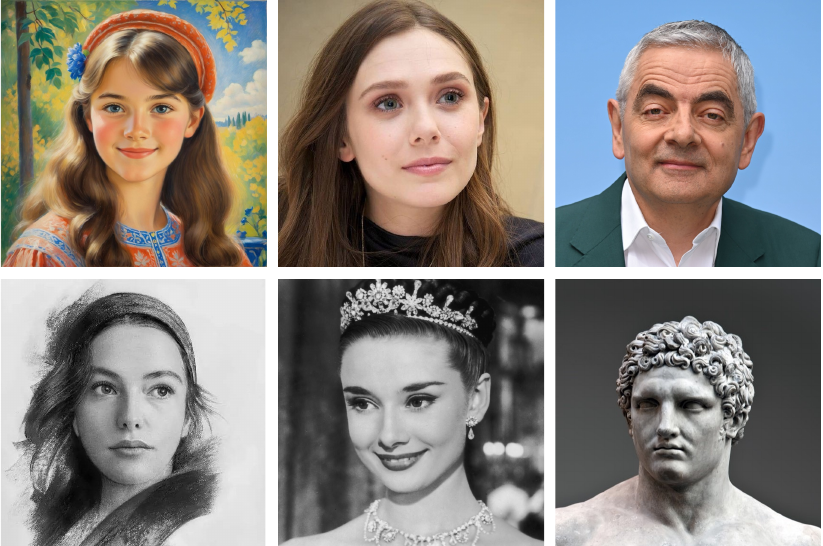}
    \caption{Talking faces in test dataset.}
    \label{fig:3}
    \end{figure}

    To thoroughly validate the effectiveness of our proposed model, JoyHallo, and ensure fair comparisons, we selected test datasets with various artistic styles that were not involved in training. Figure \ref{fig:3} illustrates some of the test datasets, highlighting their diversity. Additionally, to demonstrate JoyHallo's robust capabilities in generating both Mandarin and English videos, we conducted evaluations in both language contexts.

    \begin{table*}[!hb]
    \centering
    \renewcommand\arraystretch{1.3}
    \setlength{\tabcolsep}{1.5mm}
    \caption{Comparison of video quality in Mandarin.}    
    \begin{tabular}{cccccccc}
        \toprule
        Model     & IQA $\uparrow$  & VQA $\uparrow$  & Sync-C $\uparrow$ & Sync-D $\downarrow$ & Smooth $\uparrow$ & Subject $\uparrow$ & Background $\uparrow$ \\ \hline
        Hallo\cite{xu2024hallo}     & \textbf{0.7865} & 0.8563          & 5.7420            & \textbf{13.8140}    & 0.9924            & 0.9855             & \textbf{0.9651} \\
        JoyHallo  & 0.7781          & \textbf{0.8566} & \textbf{6.1596}   & 14.2053             & \textbf{0.9925}   & \textbf{0.9864}    & 0.9627 \\ 
        \bottomrule
    \end{tabular}
    \label{tab:1}
    \end{table*}

    \begin{table*}[!hb]
    \centering
    \renewcommand\arraystretch{1.3}
    \setlength{\tabcolsep}{1.5mm}
    \caption{Comparison of video quality in English.}    
    \begin{tabular}{cccccccc}
        \toprule
        Model     & IQA $\uparrow$  & VQA $\uparrow$  & Sync-C $\uparrow$ & Sync-D $\downarrow$ & Smooth $\uparrow$ & Subject $\uparrow$ & Background $\uparrow$ \\ \hline
        Hallo\cite{xu2024hallo}     & \textbf{0.7779} & 0.8471          & 4.4093            & \textbf{13.2340}    & 0.9921            & 0.9814             & \textbf{0.9649} \\
        JoyHallo  & \textbf{0.7779}          & \textbf{0.8537} & \textbf{4.7658}   & 13.3617             & \textbf{0.9922}   & \textbf{0.9838}    & 0.9622 \\ 
        \bottomrule
    \end{tabular}
    \label{tab:2}
    \end{table*}

    Table \ref{tab:1} presents the specific results of the video quality evaluation for Mandarin, while Table \ref{tab:2} provides the results for English. From these tables, it is evident that the generated videos exhibit superior performance in terms of video quality, lip synchronization error confidence, video smoothness, and subject consistency, indicating high video quality, lip synchronization, and subject consistency. Additionally, JoyHallo demonstrates excellent cross-language generation capabilities, performing well in both Mandarin and English video generation. However, it is observed that JoyHallo's performance slightly diminishes in image quality and background consistency, likely due to the relatively simple backgrounds in jdh-Hallo. In future tasks, post-processing of the generated videos could help mitigate background effects.

    Table \ref{tab:3} shows the comparison of memory usage and cost time during inference. The inference arguments consisted of 512x512 image size, 16-frame audio length, and a step count of 40. From Table \ref{tab:3}, it is clear that compared to the Hallo model, JoyHallo reduced memory requirements by 2.5\% and inference time by 14.3\%. The model's inference efficiency significantly improved, which is crucial for the practical deployment of such tasks.

    \begin{table}[!htb]
    \centering
    \caption{Comparison of memory usage and inference time.}
    \begin{tabular}{cccc}
    \toprule
        ~ & JoyHallo & Hallo & Improvement \\ \hline
        GPU Memory & 19049m & 19547m & \textbf{2.5\%} \\ 
        Inference Time & 24s & 28s & \textbf{14.3\%} \\ 
    \bottomrule
    \end{tabular}
    \label{tab:3}
    \end{table}

\section{Conclusions}
    In this study, we address the challenges of audio-driven video generation in Mandarin, a task complicated by the language's intricate lip movements and the scarcity of high-quality datasets. By collecting 29 hours of diverse Mandarin speech video data from JD Health International Inc. employees, we created the jdh-Hallo dataset, which encompasses a wide range of ages and speaking styles. The JoyHallo model, adapted for Mandarin using the Chinese wav2vec2 model, incorporates a semi-decoupled structure with a Cross Attention mechanism to enhance lip-sync accuracy and overall video quality.

    Our experiments demonstrate that JoyHallo significantly improves video quality, motion smoothness, and subject consistency while maintaining competitive performance in image quality and background consistency. The model also achieves a notable 14.3\% reduction in inference time and a 2.5\% decrease in memory usage compared to the Hallo model, highlighting its efficiency and potential for practical deployment. Furthermore, JoyHallo retains strong cross-language generation capabilities, effectively generating videos in both Mandarin and English. This work represents a substantial advancement in the field, offering a robust solution to the unique challenges of Mandarin video generation.

{\small
\bibliographystyle{unsrt}
\bibliography{references}
}

\end{document}